\documentclass[10pt,twocolumn,letterpaper]{article}

\usepackage[pagenumbers]{iccv} 

\usepackage{multirow}
\usepackage{colortbl}
\definecolor{iccvblue}{rgb}{0.21,0.49,0.74}
\usepackage[pagebackref,breaklinks,colorlinks,allcolors=iccvblue]{hyperref}

\usepackage{xcolor}
\definecolor{mygray}{RGB}{242, 242, 242}
\def\paperID{8494} 
\def\confName{ICCV}
\def\confYear{2025}

\title{Latent Posterior-Mean Rectified Flow for Higher-Fidelity\\ Perceptual Face Restoration}
	
\author{
    Xin Luo, Menglin Zhang, Yunwei Lan, Tianyu Zhang, Rui Li, Chang Liu, Dong Liu\\
    University of Science and Technology of China, Hefei, China\\
    {\tt\small xinluo@mail.ustc.edu.cn, \tt\small dongeliu@ustc.edu.cn}
    \\\small\textbf{\url{https://github.com/Luciennnnnnn/Latent-PMRF}}
}

\begin{document}
\maketitle
\begin{abstract}
    The Perception-Distortion tradeoff~(PD-tradeoff) theory suggests that face restoration algorithms must balance perceptual quality and fidelity. To achieve minimal distortion while maintaining perfect perceptual quality, Posterior-Mean Rectified Flow~(PMRF) proposes a flow based approach where source distribution is minimum distortion estimations. Although PMRF is shown to be effective, its pixel-space modeling approach limits its ability to align with human perception, where human perception is defined as how humans distinguish between two image distributions. In this work, we propose \textbf{Latent-PMRF}, which reformulates PMRF in the latent space of a variational autoencoder~(VAE), facilitating better alignment with human perception during optimization. By defining the source distribution on latent representations of minimum distortion estimation, we bound the minimum distortion by the VAE's reconstruction error. Moreover, we reveal the design of VAE is crucial, and our proposed \textbf{Sim-VAE} significantly outperforms existing VAEs in both reconstruction and restoration. Extensive experiments on blind face restoration demonstrate the superiority of Latent-PMRF, offering an improved PD-tradeoff compared to existing methods, along with remarkable convergence efficiency, achieving a $5.79\times$ speedup over PMRF in terms of FID. Our code will be available as open-source.

\end{abstract}    
\section{Introduction}
\label{sec:intro}

\begin{figure}[t]
  \setlength{\abovecaptionskip}{3pt}
  \centering
  \includegraphics[width=\linewidth]{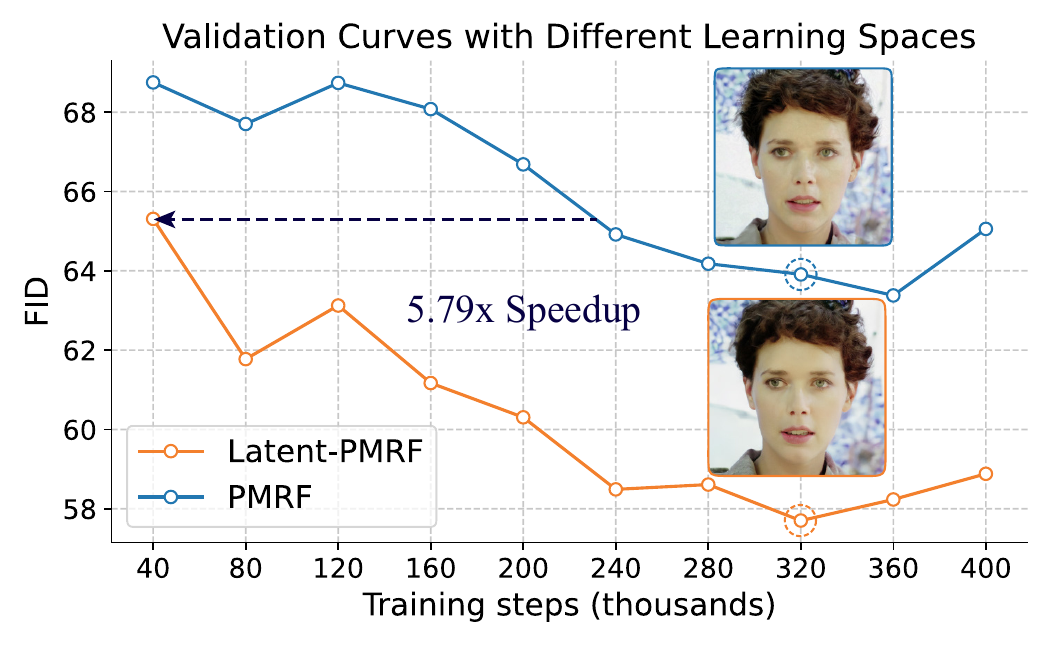}
  \caption{Illustration of perception optimization efficiency in latent space. We train PMRF and Latent-PMRF with the same compute budget. For VAEs with perceptual compression capabilities, differences in their latent space align better with human perception than those in pixel space, making latent space modeling more effective for perception optimization. Validation curves demonstrate the superior perceptual quality achieved by Latent-PMRF, with a $5.79\times$ speedup over PMRF in terms of FID.}
  \label{fig:teaser}
\end{figure}

Face images are among the most common types of images, yet they often suffer from complex degradations during formation, recording, processing, and transmission~\cite{wang2022survey}. Typical degradations, such as blur~\cite{zhang2022deep}, noise~\cite{elad2023image}, downsampling~\cite{dong2014learning,luo2023effectiveness,liang2021swinir}, and JPEG compression~\cite{jiang2021towards}, can significantly degrade visual quality. Perceptual face restoration aims to recover high-quality, visually pleasing face images from degraded inputs. The key challenge lies in enhancing perceptual quality while maintaining fidelity. Recent studies show that generative models, particularly diffusion models~\cite{sohl2015deep,ho2020denoising,li2022srdiff,wang2024exploiting} and flow matching models~\cite{zhu2024flowie,ohayon2025posteriormean}, offer strong solutions for perceptual quality by modeling the distribution of natural images. Although such posterior modeling approaches can achieve perfect perceptual quality in theory, they do not guarantee minimal distortion under perfect perceptual quality constrain~\cite{blau2018perception,freirich2021theory,ohayon2025posteriormean}. To minimize distortion, Posterior-Mean Rectified Flow~(PMRF)~\cite{ohayon2025posteriormean} transports minimum distortion estimation to the target distribution using a rectified flow model. This approach can theoretically achieve minimal distortion~\cite{freirich2021theory,ohayon2025posteriormean} under perfect perceptual quality constrain.

In this work, we challenge the necessity of constructing PMRF in the pixel space. While perceptual quality is formally defined as the statistical distance between the distributions of reconstructed and original images~\cite{blau2018perception}, researchers have found that distances in feature space better correlate with human perception~\cite{heusel2017gans,zhang2018unreasonable,szegedy2015going,sauer2021projected,kumari2022ensembling}. For instance, the most commonly used metric for evaluating image generation models is Fréchet Inception Distance~(FID)~\cite{heusel2017gans}, which measuring distribution difference within the feature space of the InceptionNet~\cite{szegedy2015going}. Additionally, many Generative Adversarial Networks~(GANs)~\cite{goodfellow2020generative} define discriminators in the feature spaces of pre-trained networks, such as EfficientNet~\cite{sauer2021projected} and CLIP~\cite{kumari2022ensembling}. These findings suggest that measuring distribution distances in feature space is an effective approach. Motivated by this, we propose reformulating PMRF in the latent space of a variational autoencoder~(VAE)~\cite{kingma2014auto}, where perceptual quality can be optimized more efficiently, as shown in Figure~\ref{fig:teaser}.

While the idea appears straightforward, its optimality in terms of distortion requires careful analysis. Analogous to PMRF, we consider two distinct source distributions: (1) the posterior mean of latent representations, and (2) the latent representations of posterior mean. We show that the second approach offers several advantages and is preferable. Most notably, it achieves minimal distortion bounded by the VAE's reconstruction error, which is not guaranteed by the first approach.

Overall, our \textbf{Latent-PMRF} can be understood as a rectified flow model~\cite{liu2023flow} in latent space, where the source distribution consists of the latent representations of the posterior mean and the target distribution consists of the latent representations of high-quality~(HQ) images. While extensive research has explored latent space models for restoration tasks~\cite{wang2024exploiting,yue2023resshift,zhu2024flowie,lin2024diffbir,gu2022vqfr,liu2023codeformer,wang2022restoreformer}, a fundamental question remains: are the commonly used VAEs sufficient for image restoration? We reveal that the VAEs employed in Stable Diffusion~(SD)~\cite{rombach2022high}, SDXL~\cite{podell2024sdxl}, and FLUX~\cite{esser2024scaling} are suboptimal for this task, as shown in Table~\ref{tab:compare_vae}. Unlike image generation, where increasing latent dimensionality often complicates optimization, restoration tasks benefit from a more informative latent space, as it reduces reconstruction error and thus lowers the minimal distortion bound.

To address this, we propose \textbf{Sim-VAE}, a simplified variant of SD-VAE, incorporating loss enhancements and architectural improvements that significantly improve both the VAE’s reconstruction ability and the restoration performance of the final model. Our contributions are summarized as follows:
\begin{table}[t]
  \centering
  \caption{\textbf{Comparison of VAEs} in CelebA-Test~\cite{wang2021towards}. We evaluate the reconstruction performance of various VAEs and their effectiveness as latent spaces for Latent-PMRF. Notably, our Sim-VAE demonstrates significantly improved reconstruction capabilities and enhances the performance of Latent-PMRF in restoration.}
  \resizebox{0.99\linewidth}{11mm}{
    \begin{tabular}{lcccccc}
      \toprule
      \multirow{2}{*}{VAE} & \multicolumn{3}{c}{Reconstruction} & \multicolumn{3}{c}{Restoration} \\
       & \multicolumn{1}{c}{PSNR$\uparrow$} & \multicolumn{1}{c}{LPIPS$\downarrow$} & \multicolumn{1}{c}{FID$\downarrow$} & \multicolumn{1}{c}{PSNR$\uparrow$} & \multicolumn{1}{c}{LPIPS$\downarrow$} & \multicolumn{1}{c}{FID$\downarrow$} \\
      \midrule
      SD1.5-VAE f8c4 & 30.463 & 0.044 & 6.937 & 25.875 & \underline{0.231} & 12.462 \\
      SD-XL-VAE f8c4 & 32.396 & 0.039 & 5.136 & \textbf{26.481} & 0.263 & \underline{12.247} \\
      FLUX-VAE f8c16 & \underline{38.763} & \underline{0.008} & \underline{0.611} & 26.152 & 0.245 & 15.999 \\
      \textbf{Sim-VAE} f8c32 & \textbf{42.712} & \textbf{0.007} & \textbf{0.431} & \underline{26.382} & \textbf{0.223} & \textbf{11.331} \\
      \bottomrule
    \end{tabular}
  }
  \label{tab:compare_vae}
\end{table}

\begin{itemize}
  \item Latent-PMRF achieves better alignment with human perception during optimization, resulting in a $5.79\times$ speedup over PMRF in terms of FID.
  
  \item The source distribution design of Latent-PMRF bounds the minimum distortion to the VAE's reconstruction error, and our improved Sim-VAE significantly boosts restoration performance when integrated with Latent-PMRF.
  
  \item Extensive experiments show that our Latent-PMRF achieves an improved PD-tradeoff and produces visually appealing results with high consistency to the inputs.
\end{itemize}
\section{Background}
\label{sec:background}

\subsection{Rectified Flow}
Rectified Flow~\cite{liu2023flow,lipman2023flow,albergo2023building} is a generative modeling approach that constructs a probability path $(p_t)_{0\leq t\leq 1}$ from a source distribution $p_0$ to a target distribution $p_1$. Sampling involves drawing $X_0 \sim p_0$ and solving an Ordinary Differential Equation~(ODE) defined by a velocity field $v_t$, which guides the transformation:

\begin{equation}
\frac{\mathrm{d}}{\mathrm{~d} t} \psi_t(x) = v_t\left(\psi_t(x)\right), \quad \psi_0(x) = x.
\end{equation}

The velocity field $v_t$ is parameterized by a neural network $v_t^\theta$ and trained via regression to match the conditional velocity field:

\begin{equation}
v_t\left(x_t \mid x_0, x_1\right) = x_1 - x_0,
\end{equation}

where $X_t$ follows a linear interpolation between $X_0 \sim p_0$ and $X_1 \sim p_1$. The training objective is to minimize the Conditional Flow Matching (CFM) loss:

\begin{equation}
\mathcal{L}_{\mathrm{CFM}}\left(\theta\right) = \mathbb{E}_{t, X_t, X_0, X_1} \left\| v_t^\theta(X_t) - \left(x_1 - x_0\right) \right\|^2.
\end{equation}

\subsection{Posterior-Mean Rectified Flow}

Let $y$ denote a low-quality~(LQ) image, which is a realization of a random vector $Y$ with probability density function $p_Y$, and let $x$ denote a high-quality image, which is a realization of a random vector $X$ with probability density function $p_X$. Posterior-Mean Rectified Flow (PMRF) is an image restoration framework designed to minimize distortion while preserving perceptual quality. PMRF achieves minimum distortion through two key stages:

1. \textbf{Posterior Mean Estimation}: A regression model is trained to estimate the posterior mean $\hat{x}=\mathbb{E}[X|Y=y]$ given a LQ image $y$. This initial estimation step is theoretically optimal for minimizing the expected distortion between the predicted and true high-quality images.

2. \textbf{Rectified Flow}: Subsequently, a rectified flow model transforms the posterior mean estimation to match the true high-quality data distribution. This is achieved by learning a velocity field $v_t^{\theta}(\cdot)$ that guides the transformation through time $t$, enabling the model to recover fine details and natural variations present in the true data distribution.

The synergy between posterior mean estimation and flow-based modeling enables PMRF to achieve superior performance in image restoration tasks. By combining a distortion-optimal initial estimate with learned continuous transformations, PMRF successfully reconstructs high-fidelity images that are both perceptually pleasing and faithful to the original content.

\section{Latent Posterior-Mean Rectified Flow}
In this section, we introduce \textbf{Latent Posterior-Mean Rectified Flow~(Latent-PMRF)}, which extends PMRF to operate in the latent space of a VAE. We first illustrate why operating in the latent space leads to more efficient optimization of perceptual quality. Then, we analyze the choice of source distribution to ensure minimal distortion. Finally, we present the complete training and sampling procedures for Latent-PMRF.

\begin{figure}[t]
    \setlength{\abovecaptionskip}{3pt}
    \centering
    \includegraphics[width=\linewidth]{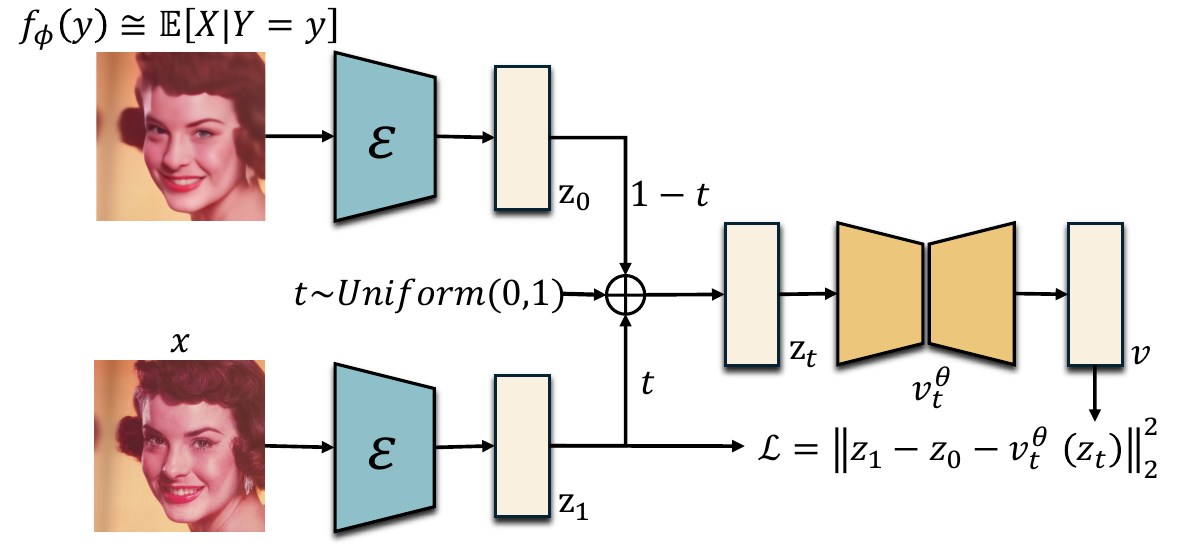}
    \caption{\textbf{Training Framework of Latent-PMRF.} We first estimate the posterior mean $\mathbb{E}\left[X|Y=y\right]$ from low-quality input $y$ using a pretrained estimator $f_{\phi}\left(\cdot\right)$. The posterior mean and the high-quality input $x$ are then encoded into latent representations $z_0$ and $z_1$. A flow network $v_{t}^{\theta}\left(\cdot\right)$ is trained to predict velocity field along their linear interpolation: $z_t = \left(1 - t\right)z_0 + t z_1$.}
    \vspace{-1.3em}
    \label{fig:framework}
\end{figure}

\subsection{Efficient Perceptual Quality Optimization}
Let $\mathcal{E}$ and $\mathcal{D}$ denote the encoder and decoder of a VAE, respectively. The high-quality latent representations is then defined as $Z = \mathcal{E}(X)$. Since rectified flow aims to transform samples from a source distribution to match a target distribution, it is natural to use $Z = \mathcal{E}(X)$ as our target distribution in the latent space.

Operating in the latent space is particularly advantageous for optimizing perceptual quality, as supported by several established practices in the field. First, perceptual metrics like LPIPS~\cite{zhang2018unreasonable}, FID~\cite{heusel2017gans} typically measure differences in feature space of pretrained networks. Second, GAN-based image generation methods~\cite{sauer2021projected,kumari2022ensembling} successfully employ feature-space discriminators for improved visual quality. Furthermore, state-of-the-art diffusion models~\cite{rombach2022high,podell2024sdxl,esser2024scaling,black2024flux} increasingly operate in VAE latent space, demonstrating the effectiveness of latent-space learning for perceptual quality optimization.

\subsection{Posterior-Mean Latent Estimation}
\label{sec:method:latent_pmrf:source_distribution}
PMRF achieves minimum distortion while preserving perceptual quality by defining the source distribution as posterior mean estimations, which are inherently optimal in terms of distortion. The choice of source distribution thus determines the lower bound of distortion that the final model can achieve.

When extending this concept to the latent space, we have two natural options for the source distribution: (1) The posterior mean of latent representations: $\mathbb{E}[\mathcal{E}(X)|Y]$, and (2) The latent representations of the posterior mean: $\mathcal{E}(\mathbb{E}[X|Y])$. We argue that option (2) is preferable as the source distribution. To demonstrate this, we analyze how closely the decoded image of source samples approximates the posterior mean $\hat{x}=\mathbb{E}[X|Y=y]$ for a given low-quality image $y$. This comparison can be formalized through the squared errors:

\begin{equation}
    \begin{aligned}
    \text{Option (1):} \quad \|\mathcal{D}(\mathbb{E}[\mathcal{E}(X)|Y=y]) - \hat{x}\|^2 \\
    \text{Option (2):} \quad \|\mathcal{D}(\mathcal{E}(\mathbb{E}[X|Y=y])) - \hat{x}\|^2
    \end{aligned}
\end{equation}

For option (2), the squared error is zero as long as the VAE achieves perfect reconstruction, \ie, $\mathcal{D}\left(\mathcal{E}\left(X\right)\right)=X$ for any possible input $X$. In contrast, option (1) imposes an additional constraint: the encoder $\mathcal{E}$ or decoder $\mathcal{D}$ must be a linear function—a condition that is not satisfied in deep neural network-based VAEs. Therefore, we adopt option (2) for our source distribution.

A key advantage of this option is that the distortion is bounded by the reconstruction capability of the VAE: better VAE reconstruction leads to lower distortion. This property partially explains why Latent-PMRF benefit from higher latent dimensions. Furthermore, this approach offers practical advantages: instead of training a dedicated model to predict the posterior mean of latents, we can utilize existing pre-trained models designed to estimate the posterior mean of images—a well-established task in the field.

\subsection{Training and Sampling Procedure}
We summarize our framework in Figure~\ref{fig:framework}. Given an input LQ image $y$, we first estimate its corresponding posterior mean $\hat{x}=\mathbb{E}\left[X|Y=y\right]$. This estimate is then encoded into latent code $z_0=\mathcal{E}\left(\hat{x}\right) \in \mathbb{R}^{d / h}$ using a pretrained VAE encoder, where $h$ is the downsampling rate of the encoder. In latent space, the objective is to estimate a probability path that transforms $z_0$ into the target latent distribution $z_1 = \mathcal{E}\left(x\right)$. The velocity network is optimized in the compact latent space, employing the same objective as vanilla flow matching with constant velocity:

\begin{equation}
    \theta^{*}=\underset{\theta}{\operatorname{argmin}}~\mathbb{E}_{t, z_t}\left[\left\|v_{t}^{\theta}\left(z_t\right) - \left(z_1-z_0\right)\right\|_2^2\right] .
\end{equation}

For the sampling process, we solve the ODE starting from the posterior mean latent $z_0$ to obtain the HQ latent $z_1$ using the Euler solver for 25 steps. The desired sample is then decoded by a pretrained VAE decoder $\mathcal{D}$ to produce the output image $\mathcal{D}(z_1)$.

\begin{figure}[t]
    \setlength{\abovecaptionskip}{3pt}
    \centering
    \includegraphics[width=\linewidth]{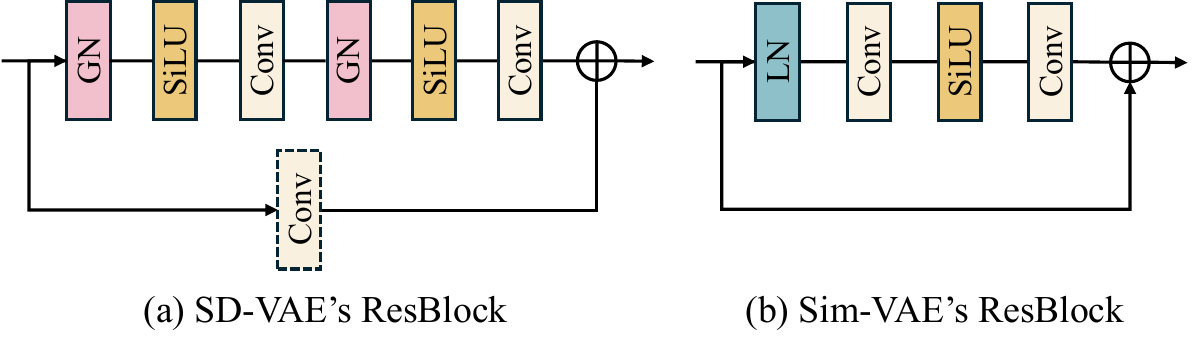}
    \caption{Comparison of ResBlock designs between SD-VAE and Sim-VAE. Sim-VAE simplifies the ResBlock architecture by removing redundant components.}
    \label{fig:simplify_resblock}
\end{figure}

\begin{figure}[t]
    \setlength{\abovecaptionskip}{3pt}
    \centering
    \begin{subfigure}[b]{0.24\linewidth}
        \setlength{\abovecaptionskip}{3pt}
        \centering
        \includegraphics[width=\linewidth]{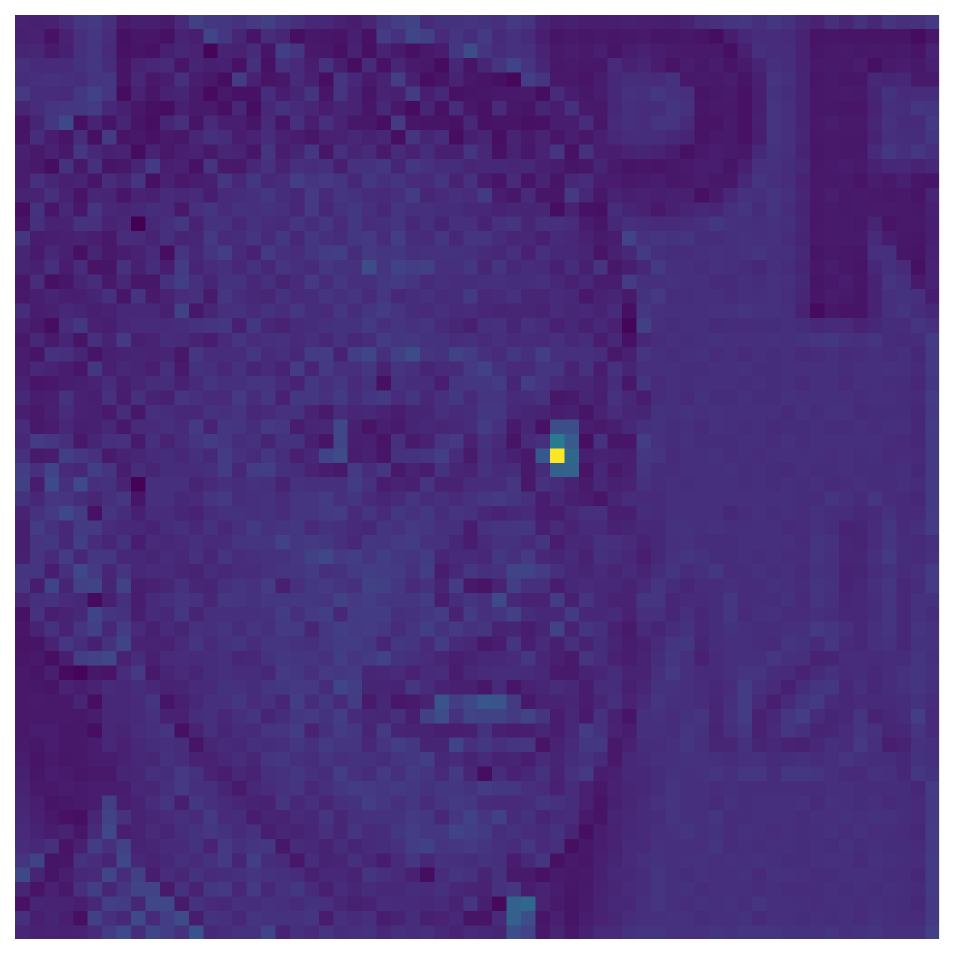}
        \caption{Group Norm}
        \label{fig:normalization_comparison:a}
    \end{subfigure}
    \begin{subfigure}[b]{0.24\linewidth}
        \setlength{\abovecaptionskip}{3pt}
        \centering
        \includegraphics[width=\linewidth]{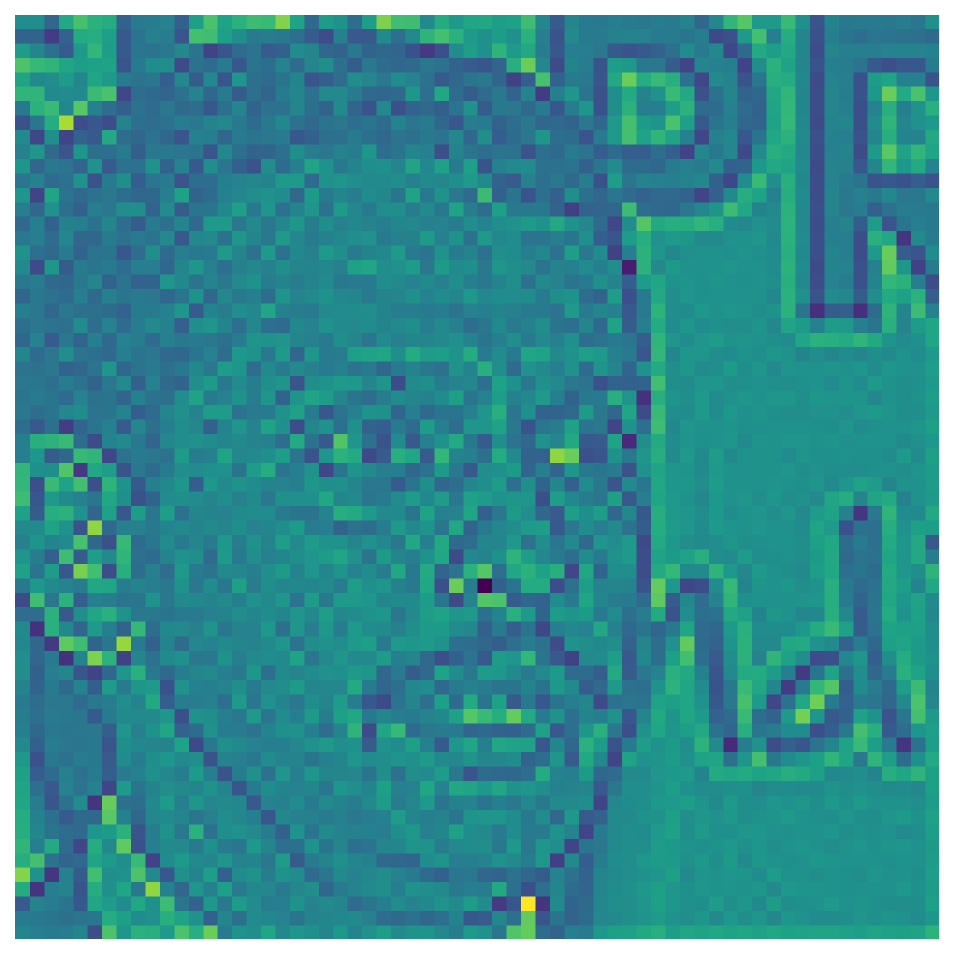}
        \caption{Layer Norm}
        \label{fig:normalization_comparison:b}
    \end{subfigure}
    \begin{subfigure}[b]{0.24\linewidth}
        \setlength{\abovecaptionskip}{3pt}
        \centering
        \includegraphics[width=\linewidth]{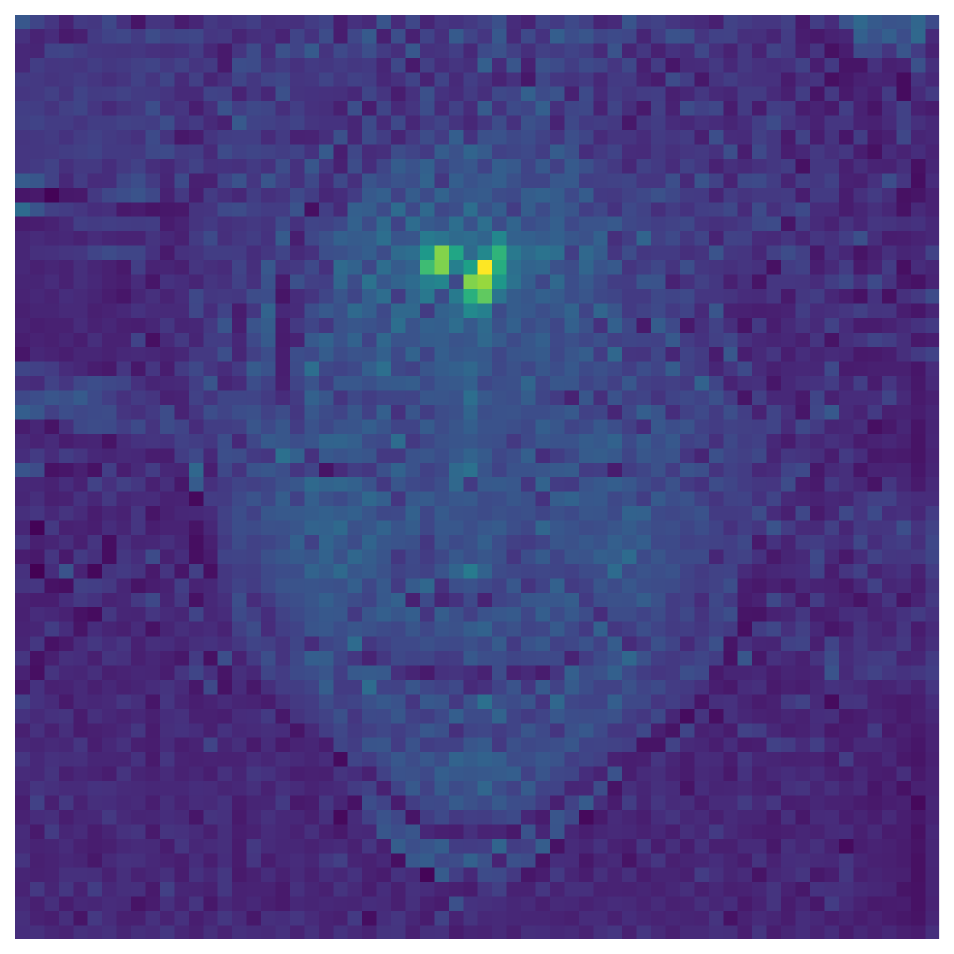}
        \caption{Group Norm}
        \label{fig:normalization_comparison:c}
    \end{subfigure}
    \begin{subfigure}[b]{0.24\linewidth}
        \setlength{\abovecaptionskip}{3pt}
        \centering
        \includegraphics[width=\linewidth]{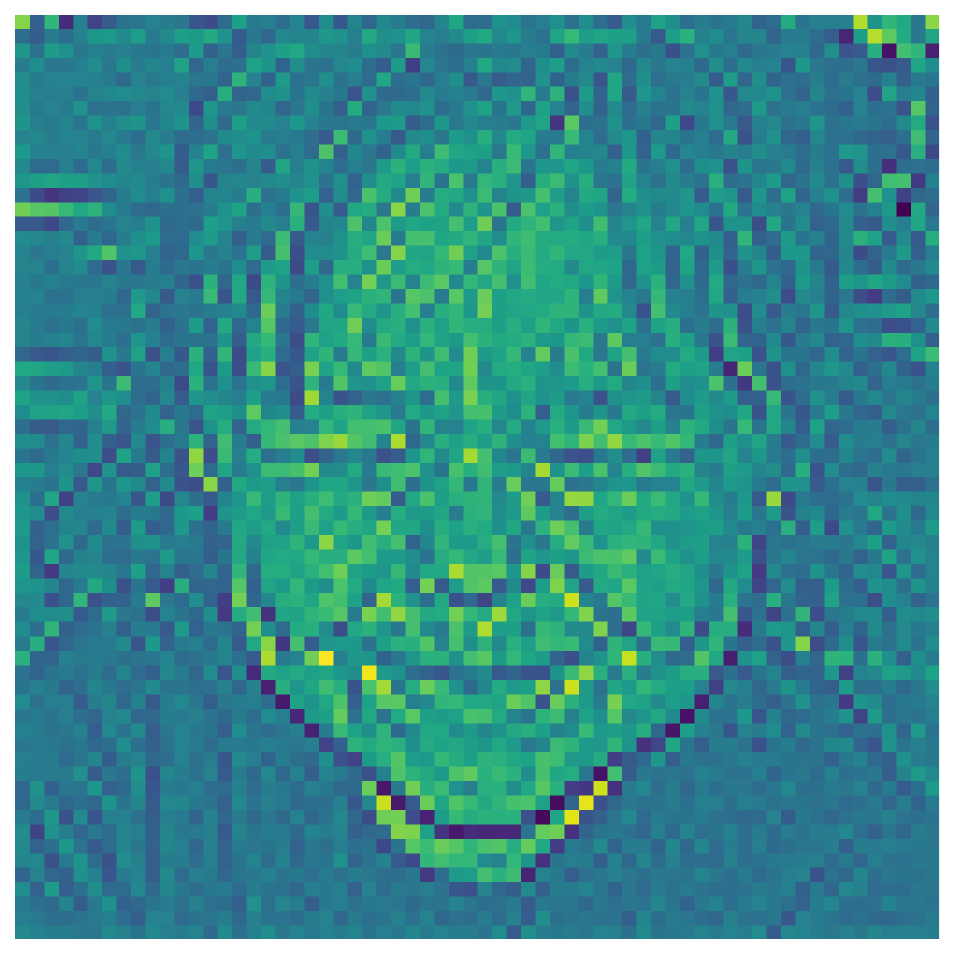}
        \caption{Layer Norm}
        \label{fig:normalization_comparison:d}
    \end{subfigure}
    \caption{Two examples of the latent representations. Using pixel-wise layer normalization instead of group normalization allows the model to learn more balanced feature maps.}
    \vspace{-1.3em}
    \label{fig:normalization_comparison}
\end{figure}

\section{Improved Variational Autoencoder}
\label{sec:method:vae}

In this section, we describe the design of \textbf{Sim-VAE}. For the Latent-PMRF model, the VAE not only defines the upper bound for restoration performance but also affects the optimization of flow model. We first outline several architectural improvements aimed at enhancing both the reconstruction ability of the VAE and the distortion lower bound of Latent-PMRF. Next, We overview our training loss, where we propose eliminating the adversarial loss when VAE is strong enough, simplifying the training procedure.

\subsection{Architecture Improvements}
\begin{figure}[t]
    \setlength{\abovecaptionskip}{3pt}
    \centering
    \begin{subfigure}[b]{0.24\linewidth}
        \setlength{\abovecaptionskip}{3pt}
        \centering
        \captionsetup{justification=centerlast}
        \includegraphics[width=0.9\linewidth]{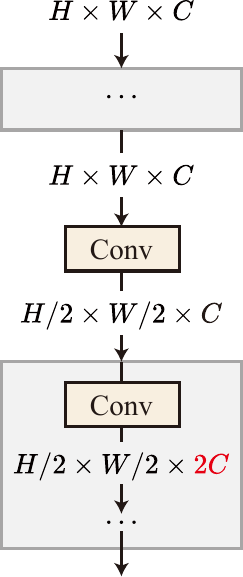}
        \caption{SD-VAE's\\Encoder}
        \label{fig:resizing_layer:a}
    \end{subfigure}
    \begin{subfigure}[b]{0.24\linewidth}
        \setlength{\abovecaptionskip}{3pt}
        \centering
        \captionsetup{justification=centerlast}
        \includegraphics[width=0.9\linewidth]{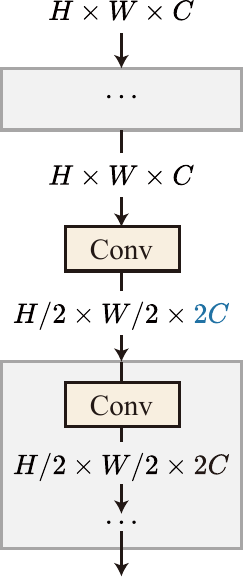}
        \caption{Sim-VAE's\\Encoder}
        \label{fig:resizing_layer:b}
    \end{subfigure}
    \begin{subfigure}[b]{0.24\linewidth}
        \setlength{\abovecaptionskip}{3pt}
        \centering
        \captionsetup{justification=centerlast}
        \includegraphics[width=0.9\linewidth]{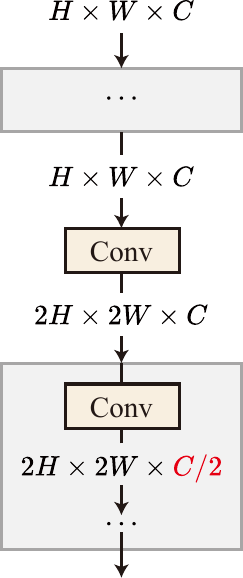}
        \caption{SD-VAE's\\Decoder}
        \label{fig:resizing_layer:c}
    \end{subfigure}
    \begin{subfigure}[b]{0.24\linewidth}
        \setlength{\abovecaptionskip}{3pt}
        \centering
        \captionsetup{justification=centerlast}
        \includegraphics[width=0.9\linewidth]{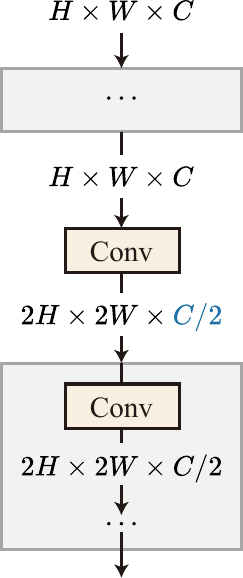}
        \caption{Sim-VAE's\\Decoder}
        \label{fig:resizing_layer:d}
    \end{subfigure}
    \caption{Illustration of the resizing layer design. Sim-VAE redistributes computation to ensure that channel dimension adjustments occur immediately with resolution changes.}
    \vspace{-1.3em}
    \label{fig:resizing_layer}
\end{figure}

Our VAE architecture builds upon the classical VQGAN~\cite{esser2021taming}, which has been widely adopted in numerous influential works~\cite{rombach2022high,podell2024sdxl,esser2024scaling,black2024flux}. We refer to this architecture as SD-VAE, reflecting its widespread adoption since Stable Diffusion. The encoder and decoder share a symmetric architecture, so we focus on describing the encoder, as the decoder follows an analogous structure in reverse.\\
\textbf{Simplified ResBlock:} Inspired by recent efficient convnet designs~\cite{liu2022convnet,chen2022simple}, we propose a simplified ResBlock~\cite{he2016deep}~(Figure~\ref{fig:simplify_resblock}) that uses only one activation function and one normalization layer per block, improving efficiency without sacrificing performance. \\
\textbf{Pixel-wise Layer Norm:} The SD-VAE has been shown to produce imbalanced feature representations, where certain regions in intermediate feature maps exhibiting disproportionately high magnitudes~\cite{sadat2024litevae,drhead2024vaeflaw}, as illustrated in Figure~\ref{fig:normalization_comparison}. While these local outliers in the feature maps serve to preserve global information~\cite{drhead2024vaeflaw}, they may complicate latent diffusion model training. Inspired by~\cite{sadat2024litevae,karras2020analyzing}, we propose replacing group normalization~\cite{wu2018group} with pixel-wise layer normalization~\cite{lei2016layer,chen2022simple}, which normalizes each spatial location independently and promotes more balanced feature representations.\\
\textbf{Removing Self-Attention in Middle Layers:}
SD-VAE uses self-attention~\cite{vaswani2017attention} in middle layers to capture global context, but this introduces a key limitation: resolution generalization issues. VAEs are usually trained on fixed low-resolution inputs, and global operators like self-attention often struggle to maintain performance across different resolutions during inference~\cite{press2022train,gao2025luminatx}. While fine-tuning on high-resolution data is a common solution~\cite{sadat2024litevae,chen2025deep}, it complicates training with additional optimization stages. To address this, we propose a simple modification: replacing self-attention with standard $3\times3$ convolutional layers, which offer better generalization across different resolutions.\\
\textbf{Redistribute Parameters between Resizing Layers:} In SD-VAE, resizing layers are responsible handling stage transitions, but the original design separates resolution changes from channel adjustments~(Figure~\ref{fig:resizing_layer:a}): resizing layers maintain channel dimensions, while later convolutional layer handle channel modifications. This creates bottlenecks during downsampling and retains inefficiently high-dimensional features during upsampling.
We propose integrating channel adjustments directly into the resizing layers—expanding channels during downsampling and reducing them during upsampling. This change improves information preservation and computational efficiency without increasing parameter count or complexity.
\subsection{Training Loss}
The training objective for autoencoders typically comprises three components~\cite{esser2021taming}: a reconstruction loss $\mathcal{L}_\text{recon}(\mathcal{D}(\mathcal{E}(x)), x)$ that measures the similarity between input and reconstructed images, a regularization term $\mathcal{L}_\text{reg}(\mathcal{E}(x))$ that constrains the latent space, and an adversarial loss~\cite{goodfellow2020generative} $\mathcal{L}_\text{adv}$ that encourages photorealistic reconstructions by discriminating between real images $x$ and their reconstructions $\mathcal{D}(\mathcal{E}(x))$. We observe that with sufficient model capacity, the adversarial loss becomes unnecessary without compromising performance. Thus, the training loss simplifies to:

\begin{equation}
    \mathcal{L}_{\text {train }}=\mathcal{L}_{\text {recon }}+\lambda_{\text {reg }} \mathcal{L}_{\text {reg }}
\end{equation}

The reconstruction loss $\mathcal{L}_{\text {recon }}$ combines $\ell_1$ distance with perceptual loss~\cite{johnson2016perceptual}, following the weighting scheme of Real-ESRGAN~\cite{wang2021real}. For regularization, we use the Kullback-Leibler~(KL) divergence as $\mathcal{L}_{\text {reg}}$, with $\lambda_{\text {reg }}$ set to $10^{-6}$ as in~\cite{rombach2022high}.

\section{Related Work}
\label{sec:related}
\textbf{Generative Models in Latent Space.} Diffusion-based generative models~\cite{ho2020denoising,dhariwal2021diffusion} achieve impressive image synthesis but are computationally expensive, particularly for high-resolution images. Latent Diffusion~\cite{rombach2022high} mitigates this by learning distributions in a pretrained VAE’s latent space, retaining only perceptually important information to enhance efficiency and scalability. Large-scale text-to-image models~\cite{rombach2022high,podell2024sdxl,esser2024scaling} follow this paradigm, with VAE design playing a crucial role. Esser~\etal~\cite{esser2024scaling} show that increasing latent channels improves performance but requires larger generative models—for instance, even FLUX (12B parameters)~\cite{black2024flux} is limited to 16 latent channels. However, our Latent-PMRF framework greatly benefits from more powerful VAE, since a stronger VAE enrich source distribution with more information, thus alleviating the burden on the restoration process.\\
\textbf{Blind Face Restoration.} Blind face restoration aims to recover high-quality facial details from images degraded by unknown and complex factors while maintaining fidelity. From a training objective perspective, existing methods mainly fall into two categories: (1) GAN-based approaches~\cite{wang2021towards,wang2022restoreformer,gu2022vqfr,liu2023codeformer} optimize a weighted combination of distortion losses~(e.g., L1, L2) and perceptual losses~(e.g., adversarial loss~\cite{goodfellow2020generative}, perceptual loss~\cite{johnson2016perceptual}), where the tradeoff between fidelity and perceptual quality is controlled by loss weighting~\cite{blau2018perception,ledig2017photo}. (2) Posterior sampling-based methods~\cite{lin2024diffbir,zhu2024flowie,yue2023resshift,yue2024difface,ohayon2025posteriormean,chen2024towards}, particularly diffusion models, model the conditional posterior distribution of HQ images given degraded inputs. While these methods theoretically ensure superior perceptual quality, they often lead to suboptimal distortion~\cite{ohayon2025posteriormean}.

PMRF~\cite{ohayon2025posteriormean} is the first approach to ensure optimal distortion under a perfect perceptual quality constraint. It first predicts the posterior mean~(minimum distortion estimation) and then transports it to the HQ image distribution. However, we argue that distribution discrepancy in pixel space does not faithfully align with human perception. To address this, we propose constructing PMRF in the latent space of a VAE, which better optimizes perceptual quality. Furthermore, we design the source distribution to preserve PMRF’s distortion-minimum properties in latent space.\\
\textbf{Concurrent works.} ELIR~\cite{cohen2025efficient} independently extends PMRF to the latent space of VAE. However, their focus is on improving testing-time efficiency via Consistency Flow Matching~\cite{yang2024consistency}, while our aim is to enhance optimization efficiency for perceptual quality. Furthermore, they use the posterior mean of latent representations as the source distribution, which, as discussed in Section~\ref{sec:method:latent_pmrf:source_distribution}, is suboptimal. This choice leads to significant fidelity degradation in their model, whereas our Latent-PMRF preserves the high fidelity of PMRF.

\section{Experiments}

\begin{figure*}[ht]
    \setlength{\abovecaptionskip}{3pt}
    \centering
    \begin{subfigure}[b]{0.24\linewidth}
        \setlength{\abovecaptionskip}{3pt}
        \centering
        \includegraphics[width=\linewidth]{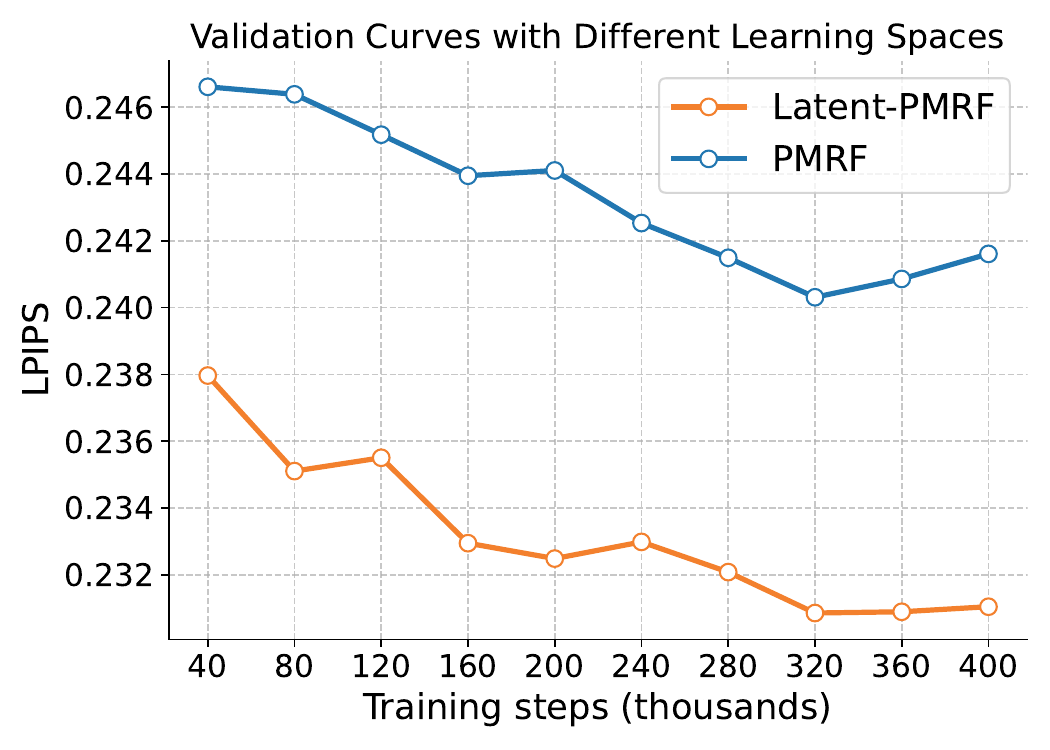}
        \caption{}
        \label{fig:convergence_efficiency:a}
    \end{subfigure}
    \begin{subfigure}[b]{0.24\linewidth}
        \setlength{\abovecaptionskip}{3pt}
        \centering
        \includegraphics[width=\linewidth]{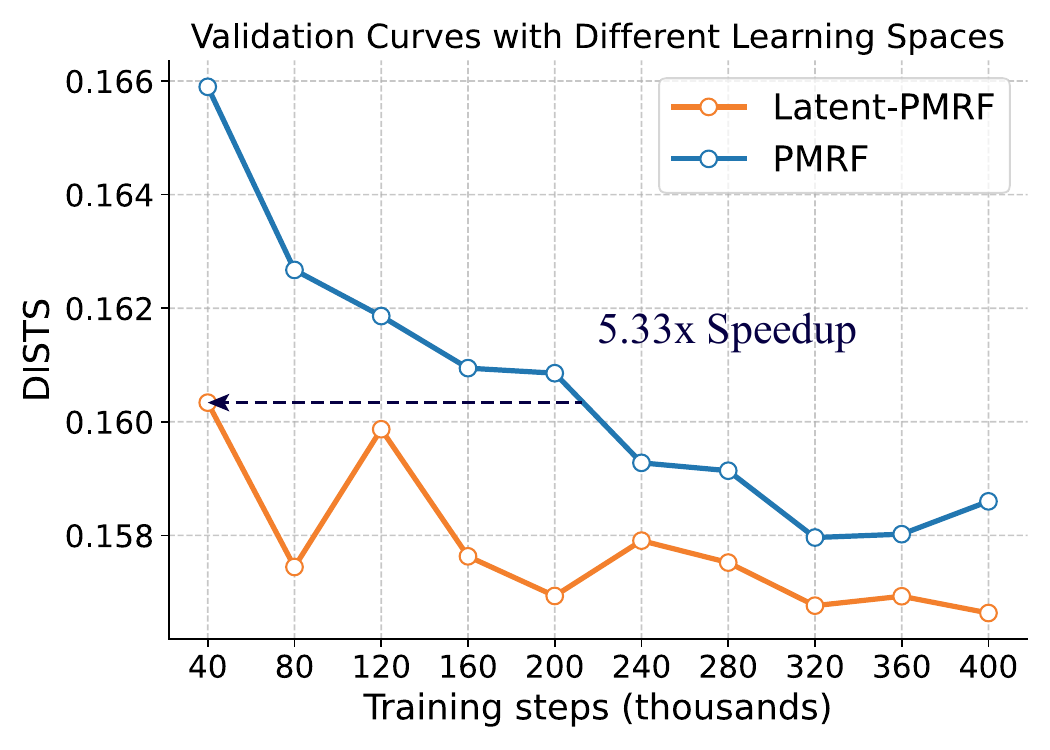}
        \caption{}
        \label{fig:convergence_efficiency:b}
    \end{subfigure}
    \begin{subfigure}[b]{0.24\linewidth}
        \setlength{\abovecaptionskip}{3pt}
        \centering
        \includegraphics[width=\linewidth]{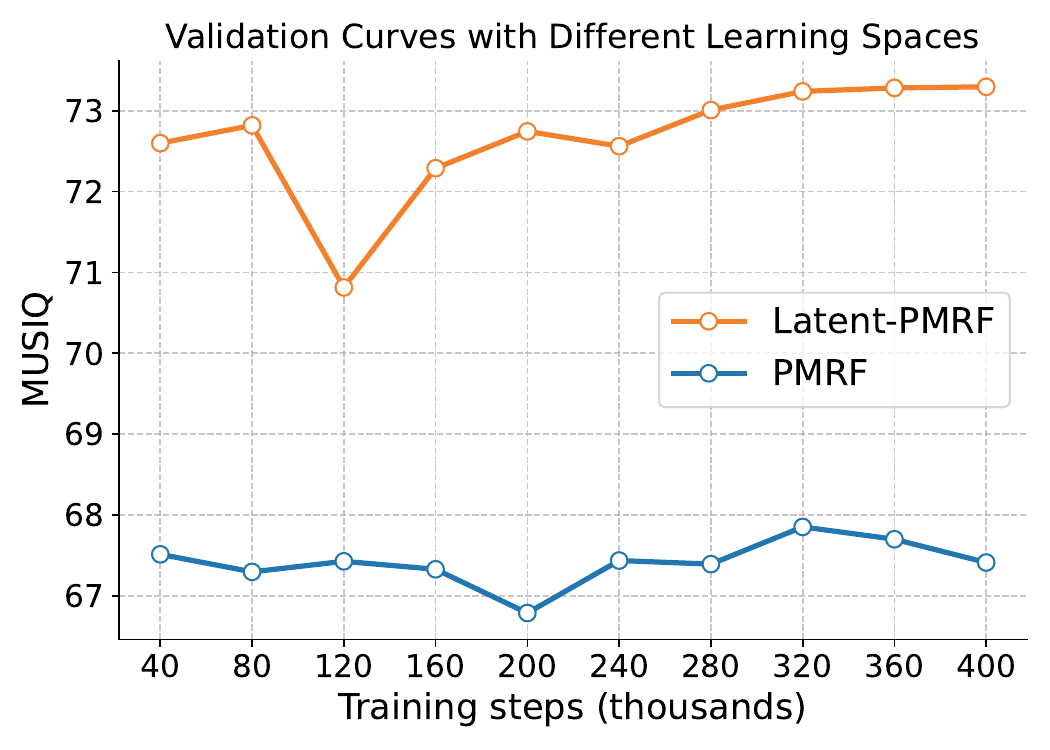}
        \caption{}
        \label{fig:convergence_efficiency:c}
    \end{subfigure}
    \begin{subfigure}[b]{0.24\linewidth}
        \setlength{\abovecaptionskip}{3pt}
        \centering
        \includegraphics[width=\linewidth]{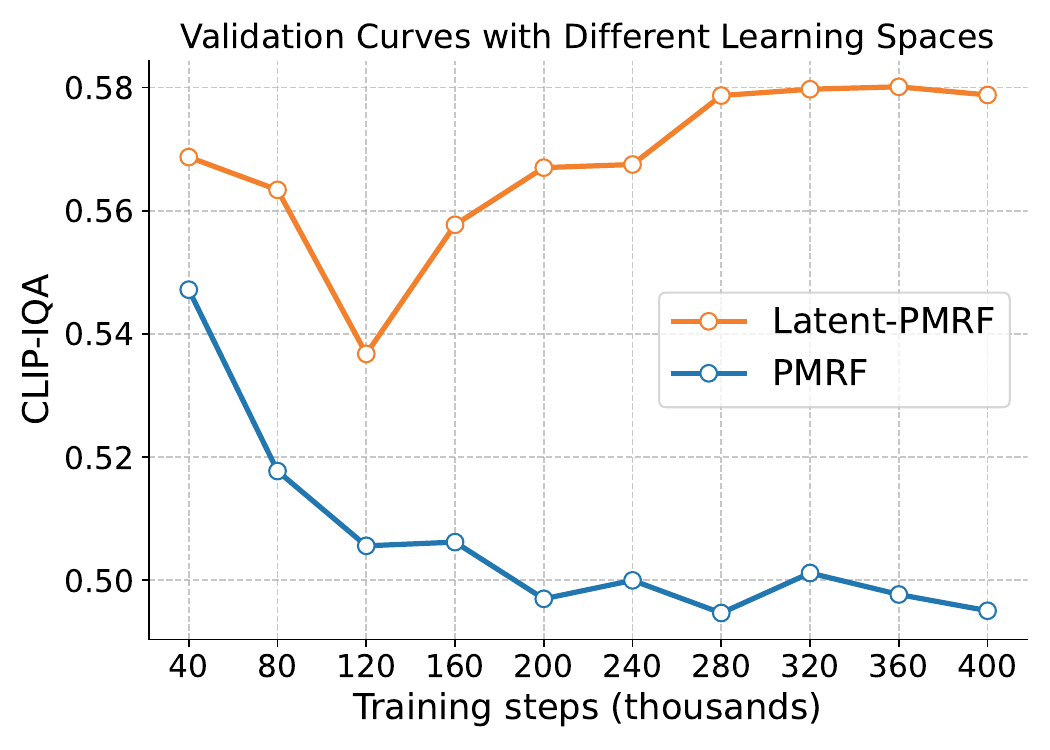}
        \caption{}
        \label{fig:convergence_efficiency:d}
    \end{subfigure}
    \caption{\textbf{Convergence Efficiency of Latent-PMRF.} We train both PMRF and Latent-PMRF using Sim-VAE for 400k iterations on FFHQ with a batch size of 64. Latent-PMRF significantly accelerates convergence, achieving a $5.33\times$ speedup in DISTS. It also outperforms PMRF in LPIPS, MUSIQ and CLIP-IQA, achieving scores that PMRF cannot achieve within training. Furthermore, Latent-PMRF demonstrates strong performance even early in training, highlighting the importance of optimizing in a well-structured latent space.}
    \vspace{-1.3em}
    \label{fig:convergence_efficiency}
\end{figure*}

\subsection{Experiment Setup}
\textbf{Datasets.} We use two primary datasets: LSDIR~\cite{li2023lsdir}, containing 84,991 high-quality natural images, and FFHQ~\cite{karras2019style}, which has 70,000 high-quality face images. For preprocessing, we crop LSDIR images into $512 \times 512$ patches and filter them using Q-Align~\cite{wu2023q} with a minimum score threshold of 3.5. FFHQ images are resized to $512 \times 512$.\\
\textbf{Implementation Details.} Sim-VAE is trained on a combination of the filtered LSDIR dataset and the first 10,000 images from FFHQ, using $256 \times 256$ image patches for 150,000 iterations with a batch size of 64. The Adam optimizer~\cite{kingma2015adam} with default parameters and a cosine learning rate schedule is used, decaying from $10^{-4}$ to $10^{-6}$ after a 500-step warmup at $10^{-5}$. We set the latent channel to 32, unless specified otherwise.

Following PMRF, We utilize the posterior mean predictor trained by~\cite{yue2024difface}, and adopt HDiT~\cite{crowson2024scalable} as velocity model of Latent-PMRF. The patch size is set to 1, and the transformer blocks are arranged as 2, 4, and 6 from high to low resolution. Depth-wise convolutions~\cite{chollet2017xception} are incorporated into both the attention and feed-forward layers. Training is performed on FFHQ for 400,000 iterations with a batch size of 64. LQ images are synthesized following ~\cite{ohayon2025posteriormean,wang2021towards}. We use the Adam optimizer~\cite{kingma2015adam} with $\beta_{1}=0.9$, $\beta_{2}=0.95$, and a fixed learning rate of $5 \times 10^{-4}$.\\
\textbf{Evaluation Metrics.}  We evaluate our methods using a range of metrics grouped into four categories:
\begin{enumerate}
    \item Reconstruction Fidelity:  PSNR and MS-SSIM~\cite{wang2003multiscale} assess reconstruction accuracy. For face restoration, we also include identity-related metrics like Deg (ArcFace embedding angle~\cite{deng2019arcface}) and landmark distance LMD~\cite{yue2024difface}.
    \item Perceptual Similarity: LPIPS~\cite{zhang2018unreasonable} and DISTS~\cite{ding2020image} measure perceptual similarity between two images.
    \item Non-Reference Metrics: CLIP-IQA~\cite{wang2023exploring}, MUSIQ~\cite{ke2021musiq} and Q-Align~\cite{wu2023q} assess image quality without ground truth.
    \item Statistical Distance: In addition to the commonly used FID~\cite{heusel2017gans} for measuring distributional differences, we also consider $\text{FID}_\text{DINOv2}$~\cite{stein2024exposing} and $\text{MMD}_\text{DINOv2}$~\cite{jayasumana2024rethinking}. These metrics improve alignment with human perception using DINOv2~\cite{oquab2023dinov2} features, while $\text{MMD}_\text{DINOv2}$ further enhances sample efficiency using Maximum Mean Discrepancy~(MMD) with an RBF kernel.
\end{enumerate}

\subsection{Convergence Efficiency of Latent-PMRF}
In this section, we demonstrate that constructing the PMRF in the latent space of Sim-VAE facilitates perception optimization, thus significantly accelerates convergence. As shown in Figure~\ref{fig:teaser} and Figure~\ref{fig:convergence_efficiency}, Latent-PMRF accelerates convergence by $5.79\times$ in terms of FID and $5.33\times$ in terms of DISTS. It also achieves significantly better LPIPS, MUSIQ and CLIP-IQA scores, outperforming standard PMRF, which fails to reach similar performance within 400k training steps. The improved convergence efficiency of Latent-PMRF allows us to achieve strong results using relatively fewer computational resources during training.

\begin{table}[t]
    \centering
    \caption{\textbf{Impact of VAE architectures} on CelebA-Test~\cite{wang2021towards}. All VAEs use 32 channels. The results show that Sim-VAE significantly outperforms SD-VAE in both reconstruction and restoration tasks. Replacing $3\times3$ convolutions with self-attention causes training instability, making results unavailable.}
    \resizebox{0.99\linewidth}{11mm}{
        \begin{tabular}{lcccccc}
            \toprule
            \multirow{2}{*}{VAE} & \multicolumn{3}{c}{Reconstruction} & \multicolumn{3}{c}{Restoration} \\
             & \multicolumn{1}{c}{PSNR$\uparrow$} & \multicolumn{1}{c}{LPIPS$\downarrow$} & \multicolumn{1}{c}{$\text{MMD}_\text{DINOv2}\downarrow$} & \multicolumn{1}{c}{PSNR$\uparrow$} & \multicolumn{1}{c}{LPIPS$\downarrow$} & \multicolumn{1}{c}{$\text{MMD}_\text{DINOv2}\downarrow$} \\
            \midrule
            \textbf{Sim-VAE} & 42.7129 & \underline{0.0073} & \textbf{0.0511} & \textbf{26.3823} & \underline{0.2236} & \textbf{0.8770} \\
            - layernorm & \textbf{43.0518} & \textbf{0.0063} & 0.0619 & 26.1698 & 0.2270 & 0.8928 \\
            - 3 × 3 conv & N/A & N/A & N/A & N/A & N/A & N/A \\
            - interpolate & \underline{42.9766} & 0.0075 & \underline{0.0556} & \underline{26.2465} & 0.2245 & \underline{0.8817} \\
            SD-VAE & 40.3979 & 0.0145 & 0.0986 & 25.2646 & \textbf{0.2224} & 0.8938 \\
            \bottomrule
        \end{tabular}
    }
    \label{tab:vae_arch}
    \vspace{-1.3em}
\end{table}

\begin{table}[t]
    \centering
    \caption{\textbf{Impact of Latent Channels} on CelebA-Test~\cite{wang2021towards}. Latent-PMRF benefits from richer latent representations, with 32 channels achieving a good balance across various metrics.}
    \resizebox{0.99\linewidth}{9mm}{
        \begin{tabular}{cccccccc}
            \toprule
            \multirow{2}{*}{Channel} & \multicolumn{3}{c}{Reconstruction} & \multicolumn{4}{c}{Restoration} \\
            & \multicolumn{1}{c}{PSNR$\uparrow$} & \multicolumn{1}{c}{LPIPS$\downarrow$} & \multicolumn{1}{c}{$\text{MMD}_\text{DINOv2}\downarrow$} & \multicolumn{1}{c}{PSNR$\uparrow$} & \multicolumn{1}{c}{LPIPS$\downarrow$} &
            \multicolumn{1}{c}{Q-Align$\uparrow$} & \multicolumn{1}{c}{$\text{MMD}_\text{DINOv2}\downarrow$} \\
            \midrule
            16 & 37.9034 & 0.0261 & 0.0966 & \underline{26.4412} & \textbf{0.2191} & 4.1006 & 0.8918 \\
            24 & 40.8142 & 0.0116 & 0.0603 & 26.3911 & 0.2251 & 4.1934 & \textbf{0.8657} \\
            \textbf{32} & \underline{42.7129} & \underline{0.0073} & \underline{0.0511} & 26.3823 & \underline{0.2236} & \underline{4.2934} & \underline{0.8770} \\
            48 & \textbf{45.0554} & \textbf{0.0033} & \textbf{0.0485} & \textbf{26.4600} & 0.2264 & \textbf{4.3055} & 0.8863 \\
            \bottomrule
        \end{tabular}
    }
    \label{tab:vae_channel}
    \vspace{-1.3em}
\end{table}

\begin{table*}[!t]
    \centering
    \caption{Quantitative comparisons on \textbf{CelebA-Test}~\cite{wang2021towards} benchmark. Our approach achieves the best PD-tradeoff, significantly reducing distortion while preserving top-tier perceptual quality. $\text{PMRF}^{\ast}$ denotes PMRF trained under the same compute budget as ours. Runtime is measured on NVIDIA A100. \#Params (M) is reported as A + B, where A represents trainable parameters and B denotes frozen parameters.}
    \resizebox{0.99\linewidth}{19mm}{
        \begin{tabular}{lccccccccccccc}
            \toprule
            Method & \multicolumn{1}{c}{PSNR$\uparrow$} & \multicolumn{1}{c}{MS-SSIM$\uparrow$} & \multicolumn{1}{c}{LPIPS$\downarrow$} & \multicolumn{1}{c}{DISTS$\downarrow$} & Deg.$\downarrow$ & LMD$\downarrow$ & \multicolumn{1}{c}{MUSIQ$\uparrow$} &
            \multicolumn{1}{c}{Q-Align$\uparrow$} & \multicolumn{1}{c}{FID$\downarrow$} & \multicolumn{1}{c}{$\text{FID}_\text{DINOv2}\downarrow$} & \multicolumn{1}{c}{$\text{MMD}_\text{DINOv2}\downarrow$} & Runtime(s) & \#Params(M) \\
            \midrule
            GFP-GAN~\cite{wang2021towards} & 24.9861 & 0.8640 & 0.2407 & 0.1720 & 34.5372 & 2.4509 & 75.2940 & \cellcolor[HTML]{F2F2F2}\underline{4.7009} & 14.8021 & 223.0202 & 1.1638 & 0.0218 & 86.4 \\
            RestoreFormer~\cite{wang2022restoreformer} & 24.6157 & 0.8443 & 0.2416 & 0.1639 & 30.9218 & 1.9389 & 73.8584 & 4.5320 & 13.4083 & 152.1276 & 1.0003 & 0.0402 & 72.7 \\
            CodeFormer~\cite{liu2023codeformer} & 25.1464 & 0.8589 & 0.2271 & 0.1700 & 35.7124 & 2.1389 & \cellcolor[HTML]{F2F2F2}\underline{75.5546} & 4.5835 & 15.3959 & 184.0517 & 1.1041 & 0.0349 & 94.1 \\
            VQFR~\cite{gu2022vqfr} & 23.7626 & 0.8278 & 0.2391 & 0.1683 & 40.9100 & 3.0436 & 73.8407 & 4.5285 & 13.6547 & 199.7024 & 1.1287 & 0.0621 & 83.5 \\
            \midrule
            DifFace~\cite{yue2024difface} & 24.7964 & 0.8233 & 0.2723 & 0.1679 & 44.1442 & 2.7230 & 69.0060 & 4.0769 & 13.5138 & 184.1844 & 1.0441 & 3.7054 & 159.7 + 15.7 \\
            DiffBIR~(v2)~\cite{lin2024diffbir} & 25.3946 & 0.8668 & 0.2654 & 0.1911 & 31.2931 & 1.5646 & \cellcolor[HTML]{F2F2F2}\textbf{76.1659} & \cellcolor[HTML]{F2F2F2}\textbf{4.8782} & 20.9181 & 156.9969 & 1.0692 & 6.3952 & 363.1 + 1319.3 \\
            ResShift~\cite{yue2023resshift} & 26.0359 & 0.8734 & 0.2464 & 0.1692 & 32.2866 & 1.8718 & 67.9784 & 4.2413 & 19.1850 & 167.3501 & 1.0534 & 0.6230 & 118.9 + 77.0 \\
            \midrule
            FlowIE~\cite{zhu2024flowie} & 24.8349 & 0.8505 & 0.2312 & 0.1585 & 32.2254 & 1.7757 & 74.1167 & 4.6108 & 17.5334 & 164.6910 & 1.0733 & 0.3877 & 398.6 + 1319.3 \\
            PMRF~\cite{ohayon2025posteriormean} & \cellcolor[HTML]{F2F2F2}\underline{26.3321} & \cellcolor[HTML]{F2F2F2}\underline{0.8740} & \cellcolor[HTML]{F2F2F2}\underline{0.2232} & \cellcolor[HTML]{F2F2F2}\textbf{0.1476} & \cellcolor[HTML]{F2F2F2}\underline{29.4504} & \cellcolor[HTML]{F2F2F2}\textbf{1.5138} & 70.4967 & 4.2227 & \cellcolor[HTML]{F2F2F2}\textbf{10.7225} & \cellcolor[HTML]{F2F2F2}\textbf{96.8752} & \cellcolor[HTML]{F2F2F2}\textbf{0.7214} & \multirow{2}{*}{0.5247} & \multirow{2}{*}{159.8 + 15.7} \\
            $\textcolor{gray}{\text{PMRF}^{\ast}}$ & \textcolor{gray}{26.6431} & \textcolor{gray}{0.8729} & \textcolor{gray}{0.2407} & \textcolor{gray}{0.1596} & \textcolor{gray}{28.9294} & \textcolor{gray}{1.3799} & \textcolor{gray}{64.9143} & \textcolor{gray}{3.7261} & \textcolor{gray}{15.1663} & \textcolor{gray}{140.6601} & \textcolor{gray}{0.8578} \\
            \textbf{Latent-PMRF~(Ours)} & \cellcolor[HTML]{F2F2F2}\textbf{26.3887} & \cellcolor[HTML]{F2F2F2}\textbf{0.8789} & \cellcolor[HTML]{F2F2F2}\textbf{0.2207} & \cellcolor[HTML]{F2F2F2}\underline{0.1576} & \cellcolor[HTML]{F2F2F2}\textbf{29.0961} & \cellcolor[HTML]{F2F2F2}\underline{1.5217} & 73.1496 & 4.3325 & \cellcolor[HTML]{F2F2F2}\underline{10.9447} & \cellcolor[HTML]{F2F2F2}\underline{110.4742} & \cellcolor[HTML]{F2F2F2}\underline{0.8108} & 0.5745 & 151.2 + 106.8 \\
            \bottomrule
        \end{tabular}
    }
    \label{tab:main_blind_face_restoration}
\end{table*}

\begin{figure*}[t]
    \setlength{\abovecaptionskip}{3pt}
    \centering
    \includegraphics[width=\linewidth]{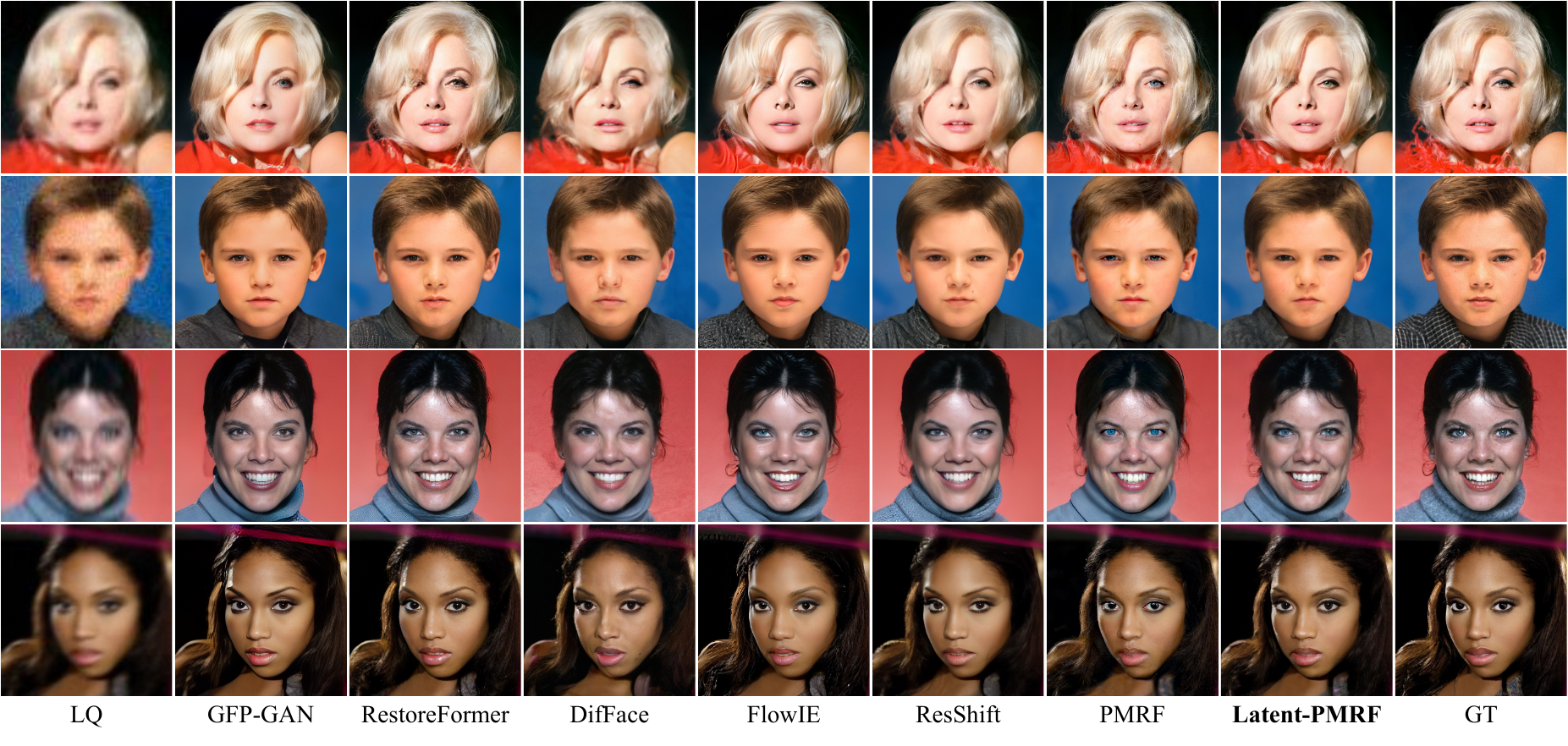}
    \caption{Qualitative comparisons on \textbf{CelebA-Test}~\cite{wang2021towards} benchmark. Our method produces visually appealing details while maintaining exceptionally high face identity preservation.}
    \vspace{-1.3em}
    \label{fig:visual_comparison_celeba}
\end{figure*}

\subsection{Improving Latent-PMRF with Better VAE}
\textbf{Effects of Architecture Design.} As illustrated in Section~\ref{sec:method:vae}, we propose a series of architectural modifications aimed at improving the learning ability of the VAE and boosting restoration of Latent-PMRF. In this section, we demonstrate the practical implications of these modifications through controlled experiments. As shown in Table~\ref{tab:vae_arch}, we progressively remove various modifications to assess their impact on the reconstruction ability of the VAE and the restoration performance of Latent-PMRF trained on it. From the second row of the table, we observe that while replacing layer normalization with group normalization improves VAE fidelity, it degrades distributional faithfulness, and more importantly, severely hampers the restoration performance of Latent-PMRF. This suggests that group normalization negatively influences the learning of smooth features. The fourth row shows that using non-optimal resizing layers leads to poorer reconstruction and, consequently, worse restoration performance. Finally, when all modifications are removed, we obtain SD-VAE, which, while achieving good LPIPS in restoration, performs poorly in all other aspects.\\
\textbf{Impact of Latent Channels.} It is well known that increasing latent channels enhances the latent space representation and improves the VAE’s reconstruction ability. However, the effect of latent channels on the restoration performance of Latent-PMRF remains unclear. As shown in Table~\ref{tab:vae_channel}, Latent-PMRF benefits from a richer latent space, with Q-Align scores consistently improving as the number of latent channels increases. We find that 32 channels strike a good balance across various metrics, so we set the default to 32.

\subsection{Comparisons with State-of-the-Art Methods}

\begin{figure*}[t]
    \setlength{\abovecaptionskip}{3pt}
    \centering
    \includegraphics[width=\linewidth]{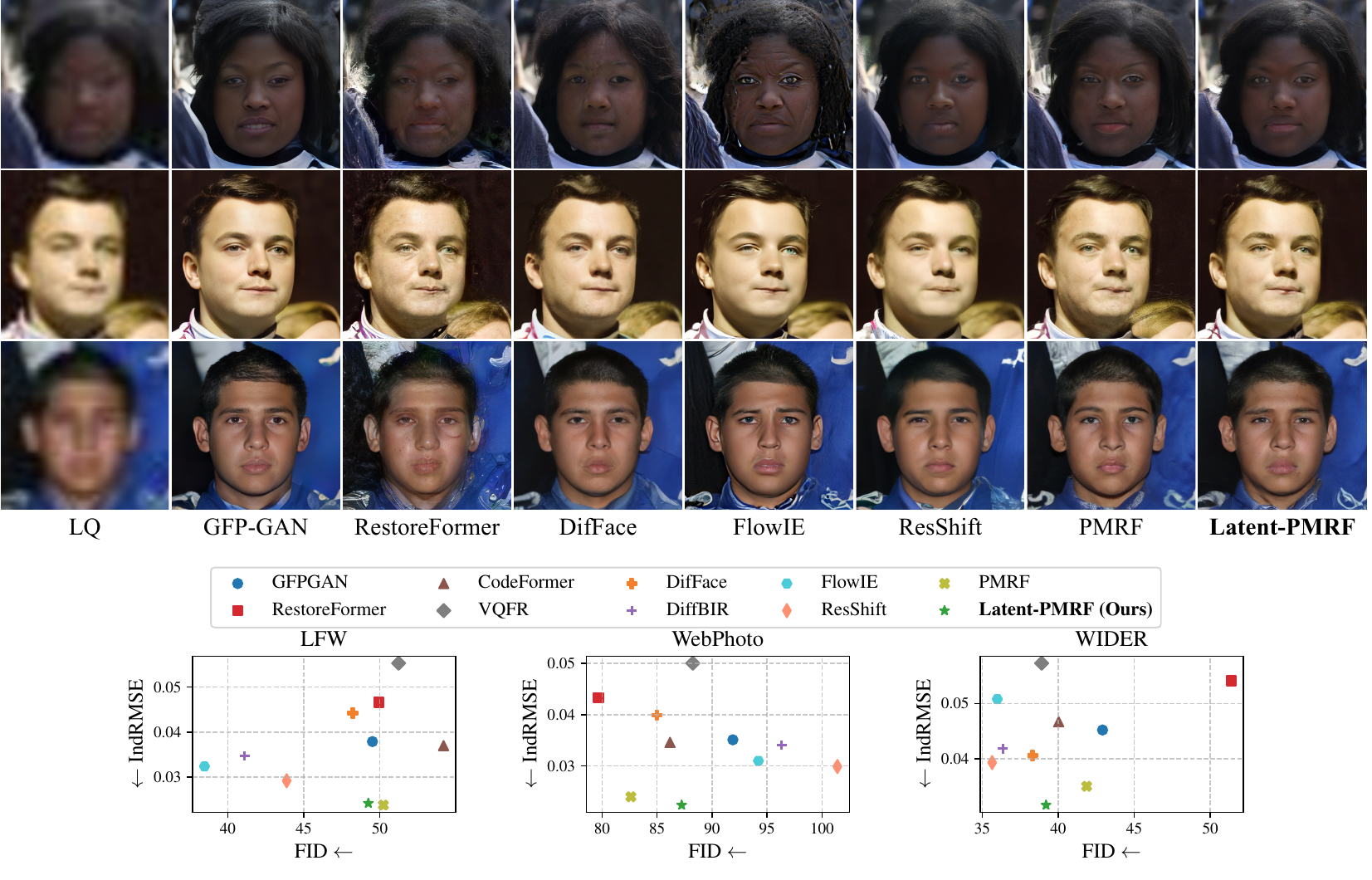}
    \caption{Comparisons on real-world datasets. Top: Qualitative results on the \textbf{WIDER-Test}~\cite{zhou2022towards} dataset. Bottom: Comparison on the "distortion"-perception plane (IndRMSE vs. FID), where IndRMSE represents the RMSE of each method~\cite{ohayon2025posteriormean}. Our method outperforms all others in IndRMSE, while achieving perceptual quality on par with the state-of-the-art.}
    \vspace{-1.3em}
    \label{fig:visual_comparison_real_world}
\end{figure*}

We primarily compare our method with PMRF~\cite{ohayon2025posteriormean}, as our goal is to construct it in the latent space. Additionally, we compare with traditional approaches such as GFP-GAN~\cite{wang2021towards}, RestoreFormer~\cite{wang2022restoreformer}, CodeFormer~\cite{liu2023codeformer}, and VQFR~\cite{gu2022vqfr}, as well as recent diffusion-based methods like DifFace~\cite{yue2024difface}, ResShift~\cite{yue2023resshift,yue2024efficient}, and DiffBIR~\cite{lin2024diffbir}. For a fair comparison, we reproduce ResShift using their official code but exclude the LPIPS loss used in their journal version. While we could incorporate this additional loss term, we omit it as it is not the focus of our work and requires computationally expensive VAE decoding during training. We also include FlowIE~\cite{zhu2024flowie}, which also utilizes flow models. Notably, both DiffBIR and FlowIE leverage facial priors from large-scale Stable Diffusion~\cite{rombach2022high}, whereas other methods use relatively smaller models.\\
\textbf{Results on Synthetic Dataset.} We evaluate our method on the CelebA-Test benchmark~\cite{wang2021towards}. As shown in Table~\ref{tab:main_blind_face_restoration}, PMRF and Latent-PMRF strike the best balance between distortion and perceptual quality. Specifically, only PMRF and Latent-PMRF achieve a PSNR above 26.3 dB and demonstrate superior face identity preservation, as evaluated by Deg. and LMD. In terms of statistical distance, PMRF, and our method learn more accurate distributions, outperforming others in FID, $\text{FID}_\text{DINOv2}$ and $\text{MMD}_\text{DINOv2}$. Notably, methods leveraging pretrained facial priors, such as GFP-GAN, DiffBIR, and FlowIE, achieve higher non-reference metric scores but tend to produce faces with lower faithfulness. In contrast, Latent-PMRF retains the high fidelity of PMRF while surpassing it in non-reference metrics. Moreover, Latent-PMRF demonstrates improved convergence properties—when the compute budget is reduced to match ours~(scaling down from a batch size 256 and 3850 epochs~\cite{ohayon2025posteriormean}), PMRF experiences a significant performance drop. Overall, Latent-PMRF not only outperforms other methods but also converges much faster than PMRF.

We also present visual results in Figure~\ref{fig:visual_comparison_celeba}. Compared to PMRF, our results generally exhibit better perceptual quality, which is reflected in the higher non-reference metrics we achieve. In contrast to other methods, which suffer from lower fidelity to the ground truth and consequently degrade face identity, our method preserves fine facial details while maintaining strong perceptual quality. \\
\textbf{Results on Real-world Datasets.} We evaluate the generalizability of Latent-PMRF on real-world datasets, including LFW~\cite{huang2008labeled,wang2021towards}, WebPhoto~\cite{wang2021towards}, and WIDER~\cite{zhou2022towards}. Since these datasets lack ground truth, we follow Ohayon~\etal~\cite{ohayon2025posteriormean} and use a pretrained posterior-mean estimator as a proxy for fidelity measurement. As shown in Figure~\ref{fig:visual_comparison_real_world}, both Latent-PMRF and PMRF significantly outperform other methods in terms of fidelity, as indicated by IndRMSE. In terms of perceptual quality, Latent-PMRF outperforms PMRF on LFW and WIDER, while maintaining comparable performance to other methods. Overall, Latent-PMRF achieves a better perception-distortion tradeoff, offering comparable perceptual quality with superior distortion reduction. Visually, RestoreFormer produces poorly structured images, and FlowIE with the Stable Diffusion backbone shows artifacts with overly sharp details. In contrast, our method generates visually appealing images that remain consistent with the input.
\section{Conclusion and Limitations}

We propose Latent-PMRF, which retains the minimal distortion property of PMRF while achieving better perceptual quality optimization. Our theoretical analysis shows that the latent representation of the posterior mean achieves a minimum distortion determined by the VAE’s reconstruction error. Based on this insight, we introduce our Sim-VAE, with a series of modifications to enhance the reconstruction capability of the VAE, leading to a notable performance boost for Latent-PMRF. Latent-PMRF demonstrates remarkable convergence efficiency, achieving a $5.79\times$ speedup over PMRF in FID convergence. Furthermore, Latent-PMRF exhibits a better PD-tradeoff compared to existing methods in blind face restoration, with improved perceptual quality compared to PMRF. Although Latent-PMRF achieves strong performance, we observe a slight decrease in test speed compared to PMRF~(see Table~\ref{tab:main_blind_face_restoration}). This is because, while the velocity prediction in the latent space is faster, the encoding and decoding processes of the VAE are inherently slow. Improving the efficiency of the VAE could be a potential area for further enhancement.

\clearpage
{
    \small
    \bibliographystyle{ieeenat_fullname}
    \bibliography{main}
}

\end{document}